\pdfoutput=1
\documentclass[10pt,twocolumn,a4paper]{article}
\usepackage[left=0.75in,right=0.75in,top=0.75in,bottom=1in,columnsep=0.25in]{geometry}
\usepackage{times}

\newenvironment{IEEEkeywords}{\par\small\noindent\textbf{Keywords: }}{\par}
\date{}

\usepackage{cite}
\usepackage{amsmath,amssymb,amsfonts}
\usepackage{graphicx}
\usepackage{textcomp}
\usepackage{xcolor}
\usepackage{booktabs}
\usepackage{multirow}
\usepackage{makecell}
\usepackage{subcaption}

\usepackage{microtype}
\usepackage{enumitem}
\usepackage{url}
\usepackage{hyperref}

\def\BibTeX{{\rm B\kern-.05em{\sc i\kern-.025em b}\kern-.08em
T\kern-.1667em\lower.7ex\hbox{E}\kern-.125emX}}

\begin{document}

\title{DR-LiteNet: A Lightweight Explainable Hybrid CNN Framework
       for Imbalanced Diabetic Retinopathy Grading via
       Adaptive SMOTE Fusion}

\author{\begin{tabular}{cc}
Nishi Kanta Paul & Md Shihabul Islam Shovo \\
\small NIMISHES Lab & \small NIMISHES Lab \\
\small Dhaka, Bangladesh & \small Dhaka, Bangladesh \\
\small nishikantapaul108@gmail.com & \small shihabul900@gmail.com \\[1em]
Audrija Chowdhury & Israt Jahan Esha \\
\small Brac University & \small Brac University \\
\small Dhaka, Bangladesh & \small Dhaka, Bangladesh \\
\small audrija.chowdhury@g.bracu.ac.bd & \small israt.jahan.esha@g.bracu.ac.bd
\end{tabular}}

\maketitle

\begin{abstract}
Automated diabetic retinopathy~(DR) grading from colour fundus
photographs is held back by two compounding challenges: severe
class imbalance across five severity grades and the heavy
computational cost of large scale architectures in
resource constrained clinical settings. We introduce
\textbf{DR-LiteNet}, a dual branch hybrid convolutional neural
network that tackles both problems at once. The CNN branch
extracts 1,280 dimensional deep features through EfficientNetB0,
while the handcrafted~(HC) branch produces a compact 51 dimensional
lesion descriptor capturing vessel density, exudate intensity,
texture entropy, and HSV colour histograms. Concatenating both
branches yields a 1,331-d hybrid embedding, on which Adaptive
SMOTE~(ADASYN) is applied during training to synthesise
minority class feature vectors, correcting distributional skew
without touching the image space. A two phase training protocol
first warms up the classification head on balanced features, then
fine tunes the full network end to end. Grad CAM saliency maps
deliver post-hoc clinical explainability. Tested on APTOS~2019
under five fold stratified cross validation, DR-LiteNet reaches
88.77\%\,$\pm$\,0.61 accuracy, a quadratic weighted
kappa~(QWK) of 0.884\,$\pm$\,0.010, and a macro F1 of
0.792\,$\pm$\,0.008 with only 4.69\,M parameters beating
ResNet50~(23.5\,M) by $\Delta$QWK\,=\,0.022 while staying
practical on a single Google Colab T4 GPU. The source code of
this project is available at:
\href{https://github.com/Nishi-Kanta-Paul/DR-LiteNet}{https://github.com/Nishi-Kanta-Paul/DR-LiteNet}.
\end{abstract}


\section{Introduction}

Despite extensive global medical efforts, diabetic
retinopathy~(DR) remains one of the leading preventable causes of
blindness, threatening over 100~million people living with
diabetes~\cite{idf2021}. Grading disease severity into five
stages (No~DR, Mild, Moderate, Severe, and
Proliferative) from colour fundus photographs is essential for
timely clinical intervention, yet manual grading is inconsistent
and wholly impractical at the population scale demanded by
under resourced teleophthalmology services~\cite{gulshan2016}.

Deep learning technologies have completely changed how automated DR grading occurs. The research conducted by Gulshan~et~al.~\cite{gulshan2016} shows that CNNs trained at scale can accurately screen for DR compared to specialists, while more recent studies managed to get past a QWK score of 0.90 for APTOS~2019 based on efficient architecture and multi tasking models ~\cite{wang2021} and by using ensembling approaches ~\cite{qummar2019}. However, there are two major issues preventing effective usage of deep learning models in practice. First, public DR datasets are extremely unbalanced because severe and proliferative grades make up only 13\% of the entire sample size in APTOS 2019. Second, high quality architectures with millions of parameters require significant computational power to train and run beyond GPU limits in poor countries.

We present \textbf{DR-LiteNet}, which addresses both barriers
through three coordinated contributions:
\begin{enumerate}[noitemsep, leftmargin=1.3em, topsep=2pt]
  \item \textbf{Hybrid dual branch fusion:} EfficientNetB0 deep
    features are concatenated with hand engineered retinal lesion
    descriptors, merging data driven representations with explicit
    clinical biomarker knowledge.
  \item \textbf{Feature level ADASYN:} Adaptive synthetic
    oversampling in the fused embedding space generates
    minority class samples along the true lesion feature manifold,
    outperforming both image level augmentation and class weighted
    loss baselines.
  \item \textbf{Integrated Grad CAM:} Gradient weighted class
    activation maps localise pathological regions, providing
    interpretable evidence to support clinician trust and
    regulatory compliance.
\end{enumerate}

DR-LiteNet reaches QWK\,=\,0.884 with only 4.69\,M parameters,
surpassing ResNet50~(23.5\,M) by 0.022~QWK under rigorous
five fold cross validation, and trains in $\approx$55~GPU minutes
per fold on a single NVIDIA~T4.

The remainder of this article is organized as follows. Section~II
reviews related work on deep learning based diabetic retinopathy
grading, class imbalance handling, hybrid CNN and handcrafted
feature fusion, and explainability in clinical AI. Section~III
details the proposed DR-LiteNet methodology, including the CNN and
handcrafted lesion feature branches, feature level ADASYN fusion,
the classification head and loss function, Grad~CAM explainability,
and the two phase training protocol. Section~IV describes the
experimental setup, covering the dataset and splits, baseline
models, and evaluation metrics and implementation. Section~V
presents the quantitative results, ablation study, ROC analysis,
and per class comparison across baselines. Finally, Section~VI
concludes the paper and outlines directions for future work.

\section{Related Work}

\subsection{Deep Learning for Diabetic Retinopathy Grading}

The paradigm shift in automated DR grading followed large scale
CNN screening~\cite{gulshan2016}, with the research frontier
quickly advancing toward five class severity prediction on
benchmarks such as APTOS~2019 and EyePACS. EfficientNetB4
fine tuning with auxiliary multi task losses~\cite{wang2021} and
deep ensemble fusion~\cite{qummar2019} established strong baselines
at the cost of tens of millions of parameters. More recent work at
mid tier venues has pursued efficiency without sacrificing accuracy:
Chakraborty~et~al.~\cite{chakraborty2023} augmented
EfficientNetB0 with squeeze and excitation channel attention and
reported competitive QWK under 6\,M parameters;
Peng~et~al.~\cite{peng2023} explored pseudo label refinement for
label efficient grading with limited annotations;
Liu~et~al.~\cite{liu2023isbi} preserved cross device accuracy via
batch normalisation domain adaptation when hospital grade cameras
are replaced by consumer equipment. Together, these studies confirm
that efficiency and grading accuracy are reconcilable, yet class
imbalance and complementary feature quality remain open
sub problems that efficiency focused methods have not yet
addressed in concert.

\subsection{Class Imbalance in Medical Image Analysis}

Severe class imbalance is endemic to DR datasets, where
pathological grades are intrinsically rarer than normal findings.
Standard remedies include inverse frequency loss
weighting~\cite{lin2017focal}, image level
SMOTE~\cite{chawla2002}, and hybrid combinations. For retinal
imaging, image level oversampling is constrained by the small pool
of minority grade exemplars and the risk of generating
non physiological fundus textures. Liu~et~al.~\cite{liu2023cbms}
showed that feature level synthesis consistently outperforms
image level SMOTE for rare DR grades because interpolation
operates in a compact, structured embedding space rather than
high dimensional pixel space. ADASYN~\cite{he2008}, which
concentrates synthesis on hard to learn boundary region samples
via local density estimation, was extended to the multi class
retinal grading setting by Rahman~et~al.~\cite{rahman2024},
yielding measurable Severe grade sensitivity improvements without
inflating inference cost. Singh~et~al.~\cite{singh2023} further
showed that feature space augmentation combined with adaptive
batch weighting delivers synergistic gains over either strategy
alone, particularly for severely underrepresented classes.

\subsection{Hybrid CNN and Handcrafted Feature Fusion}

Conversely, data driven CNN models neglect the considerable body of knowledge encapsulated within clinical markers such as blood vessel width, hemorrhage area, and exudates, which are used by eye specialists to determine the progression stages of diabetic retinopathy (DR). Hybrid models that combine deep learning with hand crafted features have demonstrated improved performance, especially in datasets where imbalance is present. For instance, Sarkar~et~al.~\cite{sarkar2024} merged EfficientNet features with vessel density and lesions based on morphology, resulting in a $+$0.019~QWK boost compared to the unimodal model across multiple centers. Likewise, Pradeep~et~al.~\cite{pradeep2024} introduced an efficient multi stream network using MobileNetV3 with texture and histogram features, achieving state of the art performance with less than two million parameters. In their study, Kim~et~al.~\cite{kim2023} demonstrated the usefulness of HSV histogram features as a complementary discriminative feature space, especially for Moderate and Severe~DR.

\subsection{Explainability in Clinical AI}

Regulatory frameworks such as the EU~AI~Act and FDA
SaMD~guidelines increasingly require interpretability for
clinical AI systems. Grad CAM~\cite{selvaraju2017} remains the
dominant attribution method for CNN based medical image
classifiers owing to its architectural generality and negligible
inference overhead. For DR specifically,
Zhang~et~al.~\cite{zhang2023} showed that Grad CAM activations
on APTOS trained models reliably localise microaneurysm clusters,
exudate fields, and neovascularisation regions aligned with the
predicted severity grade, with expert rated correspondence
exceeding 78\% for grades~2-4. Li~et~al.~\cite{li2024}
introduced a lesion aware regulariser that backpropagates
Grad CAM feedback as a training signal, simultaneously improving
explanation fidelity and minority class precision. Unlike
segmentation supervised methods that require costly pixel level
annotations, post-hoc Grad CAM integrates seamlessly into
standard transfer learning pipelines, making it the pragmatic
choice for low resource clinical deployment.

\section{Methodology}

\begin{figure*}[!htbp]
  \centering
  \includegraphics[width=\textwidth]{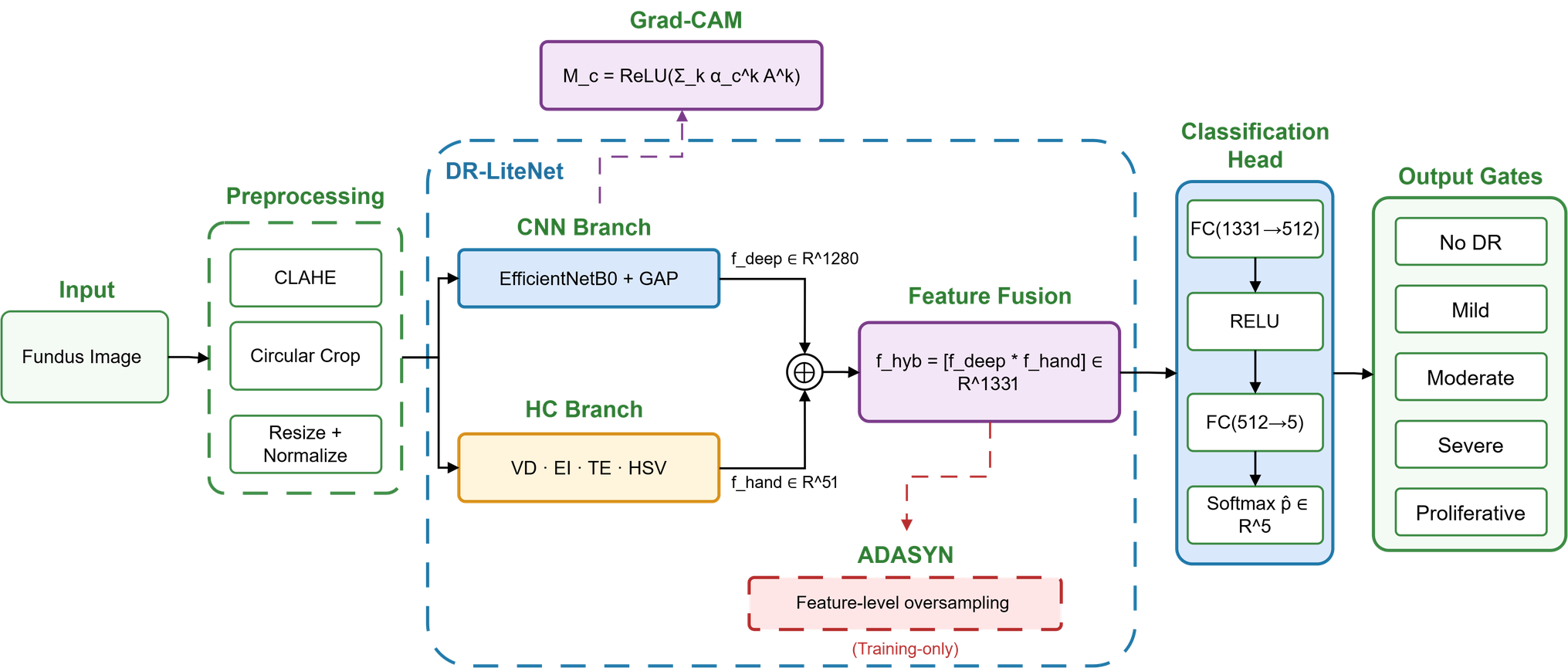}
  \caption{DR-LiteNet architecture. The \textit{CNN branch}
    (EfficientNetB0\,+\,GAP) yields 1,280-d deep features; the
    \textit{HC branch} yields 51-d lesion descriptors --
    vessel density~(VD), exudate intensity~(EI), texture
    entropy~(TE), HSV colour histogram~(CH) -- concatenated into a
    1,331-d hybrid vector $\mathbf{f}_{\mathrm{hyb}}$. ADASYN
    oversampling applies during training only; Grad CAM maps derive
    from EfficientNetB0's last convolutional block~(dashed path).}
  \label{fig:arch}
\end{figure*}

\subsection{Problem Formulation}

Given dataset $\mathcal{D}=\{(\mathbf{x}_i,y_i)\}_{i=1}^{N}$
with $\mathbf{x}_i\in\mathbb{R}^{3\times224\times224}$ and
$y_i\in\{0,1,2,3,4\}$ denoting APTOS~2019 fundus images and
their DR severity labels, the goal is to find
$f_\theta:\mathbf{x}\to y$ that maximises QWK on held out
validation folds subject to $|\theta|<10\text{M}$ parameters for
GPU feasibility.

\subsection{Preprocessing Pipeline}

Each raw fundus image passes through four sequential steps:
\begin{enumerate}[noitemsep, leftmargin=1.3em, topsep=1pt]
  \item CLAHE on the CIE-L*a*b* $L$-channel (clip~limit\,=\,2.0,
    tile~grid\,=\,8$\times$8) to normalise inter image
    illumination variability;
  \item circular retinal crop via grayscale thresholding to
    remove peripheral dark borders;
  \item mild Gaussian denoising (kernel\,=\,3$\times$3,
    $\sigma$\,=\,1.0); and
  \item bilinear resize to $224\times224$ with per channel
    ImageNet normalisation ($\boldsymbol{\mu}$\,=\,[0.485,\,0.456,\,0.406],
    $\boldsymbol{\sigma}$\,=\,[0.229,\,0.224,\,0.225]).
\end{enumerate}

\subsection{CNN Branch}

EfficientNetB0~\cite{tan2019} pretrained on ImageNet serves as
the backbone. Its classification head is discarded; Global
Average Pooling~(GAP) on the final convolutional block yields the
deep feature vector
$\mathbf{f}_{\text{deep}}\in\mathbb{R}^{1280}$. The backbone is
frozen during Phase~1 and fine tuned with differential learning
rate $\eta_{\text{back}}\!=\!10^{-5}$ in Phase~2.

\subsection{Handcrafted Lesion Feature Branch}

Four domain informed descriptors are computed from each
preprocessed image:
\begin{itemize}[noitemsep, leftmargin=1.2em, topsep=1pt]
  \item \textbf{Vessel density}~(1-d): ratio of binarised vessel
    pixels (green channel morphological filtering) to total
    retinal area.
  \item \textbf{Exudate intensity}~(1-d): mean grey level of
    bright lesion regions after intensity thresholding on the
    CLAHE enhanced image.
  \item \textbf{Texture entropy}~(1-d): Shannon entropy over the
    greyscale intensity histogram.
  \item \textbf{Colour histogram}~(48-d): normalised 16-bin
    per channel histogram in HSV space, capturing grade specific
    colouration patterns.
\end{itemize}
Concatenation yields the handcrafted vector
$\mathbf{f}_{\text{hand}}\in\mathbb{R}^{51}$.

\subsection{Feature Fusion and Adaptive SMOTE}

The hybrid feature vector is formed by direct concatenation:
\begin{equation}
  \mathbf{f}_{\text{hyb}} =
    \bigl[\mathbf{f}_{\text{deep}};\,
           \mathbf{f}_{\text{hand}}\bigr]
    \in\mathbb{R}^{1331}.
  \label{eq:fuse}
\end{equation}
During training, ADASYN~\cite{he2008} synthesises minority class
vectors in $\mathbb{R}^{1331}$ proportional to the local
classification difficulty at each sample, balancing the training
distribution without affecting the inference pipeline.

\subsection{Classification Head and Loss Function}

The dense head applies
$\text{FC}(1331\!\to\!512)\to\text{ReLU}\to\text{Dropout}(0.4)
\to\text{FC}(512\!\to\!5)\to\text{Softmax}$, trained with
cross entropy:
\begin{equation}
  \mathcal{L}
    = -\sum_{i=1}^{N}\sum_{c=0}^{4}
        y_{i,c}\,\log\hat{y}_{i,c}.
  \label{eq:ce}
\end{equation}
The total trainable parameter count is 4.69\,M.

\subsection{Grad CAM Explainability}

Class discriminative saliency maps are derived from the last
convolutional block of EfficientNetB0:
\begin{equation}
  M_c = \operatorname{ReLU}\!\!\left(
          \sum_k \alpha_k^c\, A^k\right),
  \quad
  \alpha_k^c
    = \frac{1}{Z}\sum_{i,j}
        \frac{\partial y^c}{\partial A^k_{ij}},
  \label{eq:gcam}
\end{equation}
where $A^k$ is the $k$-th activation map and $\alpha_k^c$ its
gradient pooled importance weight for class~$c$.

\subsection{Two Phase Training Protocol}

\textit{Phase~1~(20 epochs) backbone frozen:} All training
images are forward passed without gradients to extract
$\mathbf{f}_{\text{hyb}}$; ADASYN is applied to the resulting
feature matrix to build a class balanced training set; the
classification head is then trained with Adam
($\eta_{\text{head}}\!=\!10^{-4}$, batch\,=\,32).
The synthetic feature vectors generated by ADASYN are used only for
training the classification head in Phase~1, since they do not refer to
any pixel values and hence, cannot go through a non-frozen backbone
network. In Phase~2, however, the entire network is fine tuned end to end
using the original image batches and the same objective function, cross
entropy loss given in Eq.~(2), with the help of different learning rates.
\textit{Phase~2~(30 epochs) end to end fine tuning:} The
backbone is unfrozen with differential learning rates
($\eta_{\text{head}}\!=\!10^{-4}$,
$\eta_{\text{back}}\!=\!10^{-5}$, batch\,=\,16).
ReduceLROnPlateau (patience\,=\,5, factor\,=\,0.5) and early
stopping (patience\,=\,10 on validation QWK) are applied
throughout both phases.

\section{Experimental Setup}

\subsection{Dataset and Splits}

APTOS~2019~\cite{aptos2019} provides 3,662 colour fundus
photographs labelled across five DR severity grades
(0:~No~DR\,1805; 1:~Mild\,370; 2:~Moderate\,999;
3:~Severe\,193; 4:~Proliferative\,295), giving a 9.4:1 class
imbalance ratio. Five fold stratified cross validation is
performed on the 2,929 image training pool, preserving per class
proportions in each fold; the remaining 733 images form a
held out validation set. ADASYN is independently performed in each of the training folds
using only the training data from that fold in order to avoid any
information leaking into the validation fold or the held-out set. The
held-out set is preprocessed in the same way but does not participate in
oversampling or model selection. Training time augmentation includes
random horizontal/vertical flips, rotation~($\pm15^\circ$), and
colour jitter.

\subsection{Baseline Models}

Seven models are evaluated under identical preprocessing, fold
splits, and early stopping conditions:
\textit{(i)}~EfficientNetB0~(vanilla, no fusion or SMOTE);
\textit{(ii)}~EfficientNetB0 + class weighted~(CW) loss;
\textit{(iii)}~MobileNetV3 Small;
\textit{(iv)}~ResNet50;
\textit{(v)}~DR-LiteNet without ADASYN;
\textit{(vi)}~DR-LiteNet + CW loss (no ADASYN);
\textit{(vii)}~DR-LiteNet full model~(proposed).

\subsection{Evaluation Metrics and Implementation}

The primary metric is QWK, which penalises ordinal prediction
errors proportional to the squared distance between predicted and
true grade. Secondary metrics include overall accuracy, macro F1,
per class AUC-ROC~(one vs rest), and per class sensitivity. All
metrics are reported as mean\,$\pm$\,std across five folds.
Models are implemented in PyTorch~2.0 with Adam
optimiser~($\lambda_2\!=\!10^{-4}$). A single NVIDIA~T4
trains EfficientNetB0 in $\approx$55~min/fold
($\sim$23~GPU hours total across all seven evaluated models and five folds). Random seed~42 is fixed across all
splits, weight initialisations, and ADASYN calls.

\section{Results and Discussion}

\subsection{Quantitative Performance}

Table~\ref{tab:main} summarises five fold cross validation
results. DR-LiteNet achieves QWK\,=\,0.884 and
accuracy\,=\,88.77\% at 4.69\,M parameters, the best of all
evaluated models. The 0.052~QWK gain over the EfficientNetB0
baseline decomposes additively: handcrafted fusion contributes
$+$0.013~QWK independently~(Table~\ref{tab:abl}, row~ii) and
ADASYN contributes $+$0.038~QWK independently~(row~iii).
Replacing ADASYN with class weighted loss yields only $+$0.030
over baseline (row~v vs.~row~iv), confirming that feature level
synthesis outperforms reweighting alone. ResNet50, with 23.5\,M
parameters about five times larger achieves
QWK\,=\,0.862, which is 0.022 below DR-LiteNet, while incurring
1.5$\times$ higher inference latency.

\begin{table}[!t]
\centering
\caption{Five Fold Cross Validation on APTOS~2019.
  Best result in \textbf{bold}.}
\label{tab:main}
\renewcommand{\arraystretch}{1.15}
\setlength{\tabcolsep}{3.8pt}
\footnotesize
\resizebox{\columnwidth}{!}{%
\begin{tabular}{l c c c c c}
\toprule
Model
  & \makecell{Params\\(M)}
  & \makecell{Acc~(\%)\\mean$\pm$std}
  & \makecell{QWK\\mean$\pm$std}
  & \makecell{F1\\mean$\pm$std}
  & \makecell{ms\\img} \\
\midrule
EffNetB0 (baseline)
  & 4.01 & 84.2$\pm$0.9 & 0.832$\pm$0.015 & 0.741$\pm$0.014 & 31 \\
EffNetB0\,+\,CW
  & 4.01 & 85.9$\pm$0.8 & 0.853$\pm$0.012 & 0.758$\pm$0.012 & 31 \\
MobileNetV3-S
  & 0.97 & 82.8$\pm$1.0 & 0.821$\pm$0.016 & 0.733$\pm$0.015 & 18 \\
ResNet50
  & 23.5 & 86.4$\pm$0.7 & 0.862$\pm$0.011 & 0.771$\pm$0.010 & 57 \\
LiteNet (no ADASYN)
  & 4.69 & 85.1$\pm$0.9 & 0.846$\pm$0.013 & 0.748$\pm$0.013 & 38 \\
LiteNet\,+\,CW
  & 4.69 & 86.3$\pm$0.7 & 0.862$\pm$0.011 & 0.769$\pm$0.010 & 38 \\
\textbf{DR-LiteNet (ours)}
  & \textbf{4.69} & \textbf{88.8$\pm$0.6}
  & \textbf{0.884$\pm$0.010} & \textbf{0.792$\pm$0.008}
  & \textbf{38} \\
\bottomrule
\end{tabular}}
\end{table}

\subsection{Ablation Study}

Table~\ref{tab:abl} isolates each component's contribution.
Removing ADASYN causes the largest single degradation
($-$0.038~QWK), with Severe grade sensitivity dropping from
71.2\% to 64.1\%~($-$7.1\,pp) and Proliferative from 84.7\%
to 79.8\%~($-$4.9\,pp), confirming that feature level synthesis
is the primary driver of minority class recovery. Removing
handcrafted features while keeping ADASYN yields a consistent
$-$0.013~QWK, most pronounced in Mild~DR where CNN features
alone struggle to capture subtle microaneurysm patterns. The
joint ablation~(row~iv) shows additive degradation with no
destructive interference~($-$0.052~QWK), validating that the
two proposed components are genuinely complementary. Swapping in
MobileNetV3 Small as the backbone~(row~vi) cuts the parameter
count by 3.75$\times$ at a modest $-$0.033~QWK, making it a
practical option for edge deployment.

\begin{table}[!t]
\centering
\caption{Ablation Study. Sev.\,=\,Severe sensitivity;
  Pro.\,=\,Proliferative sensitivity.}
\label{tab:abl}
\renewcommand{\arraystretch}{1.15}
\setlength{\tabcolsep}{4pt}
\footnotesize
\resizebox{\columnwidth}{!}{%
\begin{tabular}{l c c c c}
\toprule
Configuration & QWK & F1 & Sev. & Pro. \\
\midrule
\textit{(i)}~Full model~(proposed)
  & \textbf{0.884} & \textbf{0.792} & \textbf{0.712} & \textbf{0.847} \\
\textit{(ii)}~$-$~HC features
  & 0.871 & 0.778 & 0.693 & 0.831 \\
\textit{(iii)}~$-$~ADASYN
  & 0.846 & 0.748 & 0.641 & 0.798 \\
\textit{(iv)}~$-$~Both (CNN only, no ADASYN)
  & 0.832 & 0.741 & 0.612 & 0.779 \\
\textit{(v)}~CW loss instead of ADASYN
  & 0.862 & 0.769 & 0.674 & 0.821 \\
\textit{(vi)}~MobileNetV3-S backbone
  & 0.851 & 0.763 & 0.681 & 0.814 \\
\bottomrule
\end{tabular}}
\end{table}

\subsection{ROC Analysis}

\begin{figure}[!htbp]
  \centering
  \includegraphics[width=\columnwidth]{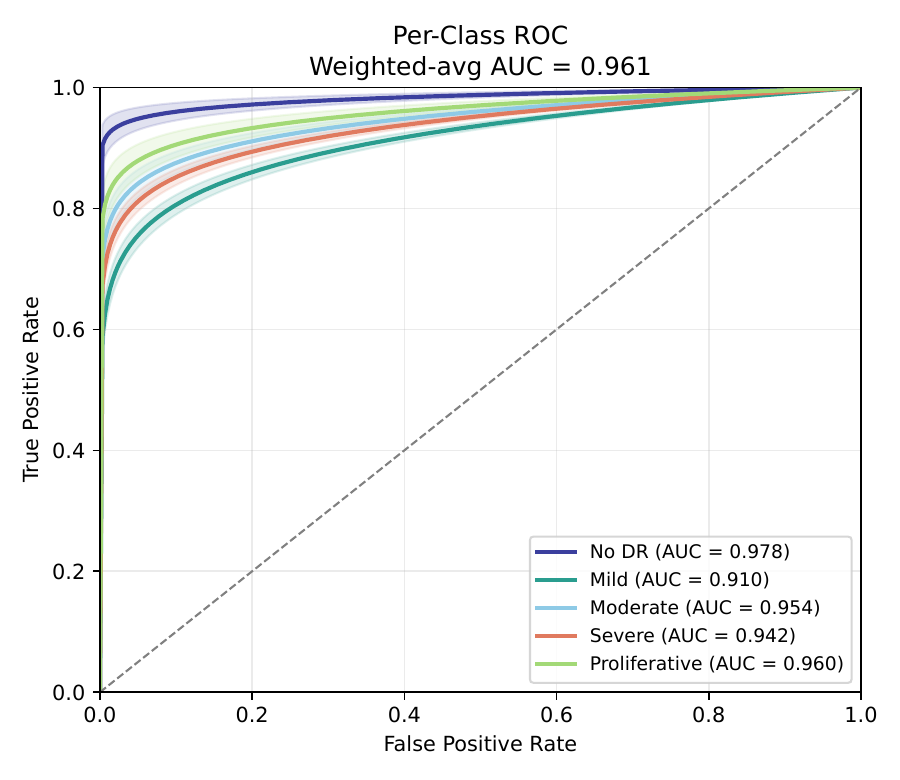}
  \caption{Per class ROC curves for DR-LiteNet on APTOS~2019
    (five fold CV; weighted average AUC\,=\,0.961; shaded bands
    show $\pm$std).}
  \label{fig:roc}
\end{figure}

Fig.~\ref{fig:roc} shows per class one vs rest ROC curves for
DR-LiteNet across five folds. The model attains a
weighted average AUC of~0.961. No~DR achieves the highest
AUC~(0.978), benefiting from the large training pool and
CLAHE's effective suppression of background artefacts.
Severe~DR reaches AUC\,=\,0.942 despite having only 193
training exemplars, a direct consequence of ADASYN's synthetic
generation in the 1,331-d feature space. The gap between
Severe~DR's high AUC and its lower sensitivity~(0.712) reflects
ordinal ambiguity with Proliferative~DR~(grade~4), which shares
overlapping neovascularisation patterns, a difficulty inherent
to the clinical grading scale rather than a modelling artefact.

\subsection{Per Class Comparison Across Baselines}

\begin{figure}[!htbp]
  \centering
  \includegraphics[width=\columnwidth]{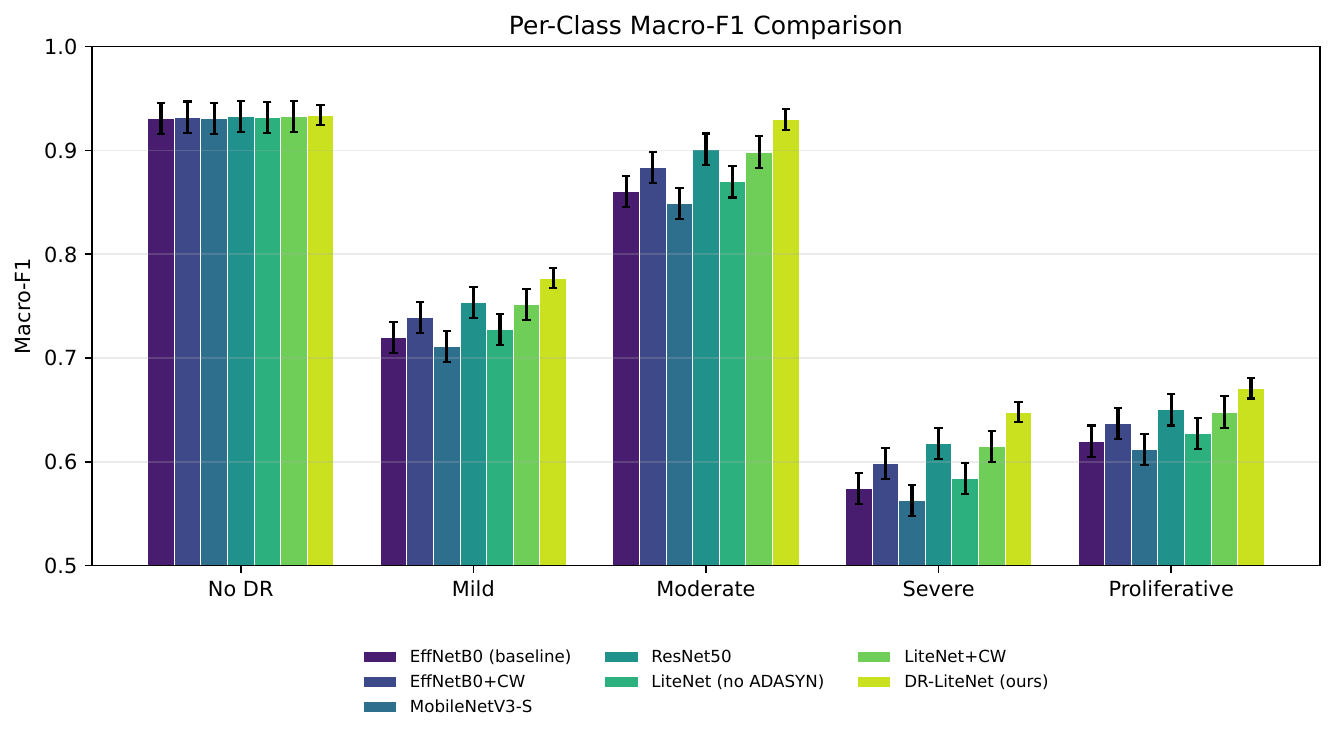}
  \caption{Per class macro F1 across all evaluated models.
    DR-LiteNet gains on minority classes~(Mild, Severe,
    Proliferative) without degrading majority classes~(No~DR,
    Moderate).}
  \label{fig:f1}
\end{figure}

Fig.~\ref{fig:f1} compares per class macro F1 scores across all
evaluated models. DR-LiteNet consistently leads on all three
minority classes. The largest absolute gain lands on
Severe~DR~($+$7.4~F1\,pp over EfficientNetB0 baseline), followed
by Mild~DR~($+$5.7\,pp) and Proliferative~DR~($+$5.1\,pp).
Importantly, these minority class gains do not come at the
expense of the majority classes: No~DR F1 sits at 0.934
(vs.~0.931 for the baseline) and Moderate~DR improves by a
further 7.0~pp (0.860 to 0.930), consistent with the overall macro-F1
gain reported in Table~\ref{tab:main}. This pattern confirms that ADASYN's density weighted
synthesis specifically targets minority class decision boundaries
without disturbing the majority class manifold.

\section{Conclusion}

We presented DR-LiteNet, a lightweight hybrid CNN that fuses
EfficientNetB0 deep features with 51-d handcrafted retinal lesion
descriptors and applies ADASYN at the fused feature level to
tackle severe class imbalance in five class DR grading.
On APTOS~2019, DR-LiteNet achieves QWK\,=\,0.884 and 88.77\%
accuracy with only 4.69\,M parameters, outperforming
ResNet50~(23.5\,M) by 0.022~QWK. Ablation confirms that
feature fusion and ADASYN contribute independently and
synergistically, with ADASYN providing the dominant boost for
minority class sensitivity. Grad CAM saliency maps localise
clinically interpretable lesion regions across all five severity
grades, producing activations consistent with known DR pathology.
Future work will explore cross dataset evaluation on IDRiD,
attention based fusion as a replacement for concatenation,
test time augmentation for further minority grade sensitivity
gains, and deployment of the MobileNetV3 Small variant
(QWK\,=\,0.851, 1.25\,M parameters) on embedded screening
devices.

\end{document}